\documentclass[sigconf]{acmart}
\usepackage{hyperref}       
\usepackage{url}            
\usepackage{booktabs}       
\usepackage{amsfonts}       
\usepackage{nicefrac}       
\usepackage{microtype}      
\usepackage{lipsum}
\usepackage{xcolor}
\usepackage{multicol}
\usepackage{amsmath}
\usepackage{ bbold }
\usepackage{caption}
\usepackage{multirow}
\usepackage{textcomp}
\usepackage{graphicx}

\copyrightyear{2021}
\acmYear{2021}
\setcopyright{acmlicensed}
\acmConference[ICMI '21 Companion]{Companion Publication of the 2021 International Conference on Multimodal Interaction}{October 18--22, 2021}{Montréal, QC, Canada}
\acmBooktitle{Companion Publication of the 2021 International Conference on Multimodal Interaction (ICMI '21 Companion), October 18--22, 2021, Montréal, QC, Canada}
\acmPrice{15.00}
\acmDOI{10.1145/3461615.3486575}
\acmISBN{978-1-4503-8471-1/21/10}
\begin{document}

%

\title{Addressing Data Scarcity in Multimodal User State Recognition by Combining Semi-Supervised and Supervised Learning}

\author{Hendric Voß}
\email{hvoss@techfak.uni-bielefeld.de}
\affiliation{%
  \institution{Social Cognitive Systems Group}
  \institution{Bielefeld University}
  \streetaddress{Universitätsstraße 25}
  \country{Germany}
}

\author{Heiko Wersing}
\email{Heiko.Wersing@honda-ri.de}
\affiliation{%
  \institution{Honda Research Institute Europe}
  \streetaddress{Carl-Legien-Str.30}
  \country{Germany}
}

\author{Stefan Kopp}
\email{skopp@techfak.uni-bielefeld.de}
\affiliation{%
  \institution{Social Cognitive Systems Group}
  \institution{Bielefeld University}
  \streetaddress{Universitätsstraße 25}
  \country{Germany}
}

\renewcommand{\shortauthors}{Voß et al.}

\begin{abstract}
Detecting mental states of human users is crucial for the development of cooperative and intelligent robots, as it enables the robot to understand the user's intentions and desires. Despite their importance, it is difficult to obtain a large amount of high quality data for training automatic recognition algorithms as the time and effort required to collect and label such data is prohibitively high. In this paper we present a multimodal machine learning approach for detecting dis-/agreement and confusion states in a human-robot interaction environment, using just a small amount of manually annotated data. We collect a data set by conducting a human-robot interaction study and develop a novel preprocessing pipeline for our machine learning approach. By combining semi-supervised and supervised architectures, we are able to achieve an average F1-score of 81.1\% for dis-/agreement detection with a small amount of labeled data and a large unlabeled data set, while simultaneously increasing the robustness of the model compared to the supervised approach.
\end{abstract}



\maketitle

\section{Introduction}
Recognizing mental states of a conversational partner is essential in human communication, as they provide important information during everyday social interaction. By recognizing signals of agreement, disagreement, and confusion during a conversation, an interlocutor can give adequate answers to a given question, make affirmative statements, and quickly identify problems during a conversation, without being dependent on explicit keywords. Although dis-/agreement and confusion is sometimes communicated verbally, many cues and indicators for these complex mental states are expressed through non-verbal behaviour \cite{Cohen2003}, like head movements or hand gestures. In everyday interactions, the human face is a particularly important source of spontaneous reactions, as it can express a wide variety of different complex mental states, like conveying interest or indicating confusion \cite{baron1996reading, Kaliouby2004}. This can be an especially important but also highly complex problem for emotionally intelligent robots \cite{Littlewort2011, Pantic2003}, and affect-sensitive human computer interaction \cite{Lang2012}, as many people convey similar reactions in very different ways.\\
In recent years, machine learning algorithms have become one of the leading methods for automatically detecting and recognizing social cues in human-robot interaction \cite{Vrigkas2017,Borges2019}. While supervised machine learning algorithms can solve an ever-increasing complexity of problems, they require a lot of accurately labeled data, which is not always available due to time and budget constraints. As the information in these small data sets are limited, the machine learning models trained on this data can be unable to function outside of narrow scenarios, due to the high noise inherent in the multimodal nature of human interaction in the wild. Semi-supervised algorithms attempt to alleviate some of these problems by training on small amounts of labeled data combined with very large unlabeled data sets \cite{Zhu2008,VanEngelen2020,Ouali2020}, but are primarily focused on computer vision \cite{Jing2019,zhai2019s4l,iscen2019label} and natural language problems \cite{Jo2019,Cho2019,Dhingra2018}. \\
Previous work on automatic dis-/agreement recognition has focused on using verbal \cite{Hillard2003, Galley2004, Hahn2006, Germesin2009} or non-verbal cues \cite{Kaliouby2004, Sheerman-Chase2009, Dunn2011}, with only a limited number of approaches combining both modalities \cite{Bousmalis2011, HosseinKhakiElifBozkurt2016}. 
For verbal cues, Wang \textit{et al.} developed an approach on conditional random fields, with which they achieved an F1-score of 57.2\% for agreement and 51.2\% for disagreement \cite{Wang2011}. Sheerman‐Chase \textit{et al.} used non-verbal cues, gathered from natural conversations with an AdaBoost classifier. Although they did not include the disagreement class, due to lack of data, they report an AUC score 0.70 for their agreement detection \cite{Sheerman-Chase2009}.
Similarly to Wang \textit{et al.}, Bousmalis \textit{et al.} developed an approach based on a hidden conditional random field algorithm that, through the combination of verbal and nonverbal cues, achieved a combined accuracy of 64.2\% for dis-/agreement detection \cite{Bousmalis2011}. 
The detection of confusion is primarily focused on EEG-based data \cite{zhou2018confusion, ni2017confused, Yang2016} or based on natural language processing with text input \cite{zeng2017learner, geller2020confused}. Using automatically recognized Action Units from the Facial Action Coding System, Borges \textit{et al.} trained an LSTM neural network on their own collected data, with which they reported an F1-score of 80.89\% for their confusion class. Semi-supervised approaches are primarily focused on computer vision and natural language processing tasks \cite{Jing2019,zhai2019s4l,Jo2019,Cho2019}, but there has been some research into using semi-supervised learning for emotion recognition. Parthasarathy \textit{et al.} trained a ladder network with an unsupervised auxiliary task to classify emotion from speech \cite{parthasarathy2020semi}. Similarly, Liang \textit{et al.} used a modified transformer architecture to classify from a multimodal data stream, consisting of audio, video and text. \\
In this work, we explore the use of unsupervised, supervised and semi-supervised learning for the identification of conversationally relevant user states in a multimodal human-robot interaction environment. We formalize user state recognition as a 4-way classification problem (agreement/disagreement/confusion/neutral) for each modality. We create a user study to gather a small labeled data set in a controlled environment and combine it with an unlabeled data set from a wide range of different scenarios, gathered from the internet. After training an unsupervised autoencoder network for audio feature extraction, we combine three different models during training for a combined semi-supervised and supervised approach. In the remainder of this paper we report on the data gathering, the learning approach, and evaluation results.

\begin{table}[ht]
 \caption{Labeled data of the three different collected data sets, in minutes}
  \centering
  \resizebox{\columnwidth}{!}{
  \begin{tabular}{lllll}
    \toprule
    Class  & audio data only & face data only & combined      \\
    \midrule
    Agreement & 5:12 & 5:35 & 4:25    \\
    Disagreement     & 6:11 & 5:52 & 5:11   \\
    Confusion     & 4:46 & 5:03 & 4:12         \\
    Neutral     & 31:46 & 31:46 & 31:46         \\
    \bottomrule
  \end{tabular}
  }
  \label{tab:table_groundtruth}
\end{table}

\begin{figure}
\centering
\includegraphics[width=6cm]{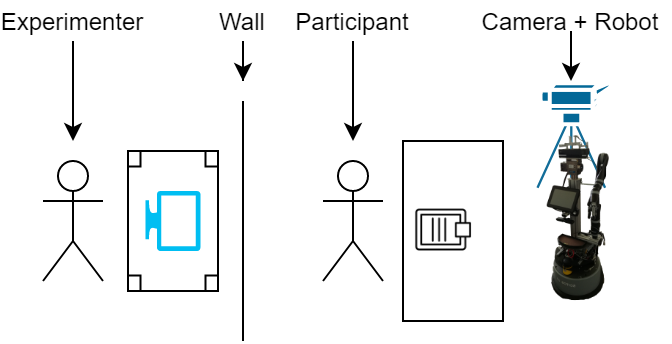}
  \captionof{figure}{The setup of our study. The participant received tasks from the experimenter through the robot, separated by a wall. All interactions were recorded by the robot as well as a backup camera.}
  \Description[Study setup]{The setup of the study, which shows the experimenter with a computer behind a wall. On the other side is the participant and the robot. The participant and the robot are separated by a table, on which the information sheet lies.}
\label{fig:experiment}
\end{figure}

\section{Data}
\label{sec:data}
Since there is limited high quality interaction data between robots and humans openly available, in which full conversations are conducted, a stand-alone study was designed, in which a participant performed different tasks with a robot as their conversational partner.

\subsection{Robot Interaction Study}
The study was conducted using a robot based on the Scitos G5 research platform, equipped with a Microsoft Kinect 2 camera and a Seeed Studio ReSpeaker 6 Microphone Array. The video data was recorded as a RGB image with a resolution of 1920x1080 at 30 FPS using the H.264 codec. The audio data was captured as a stereo signal with a bit rate of 192 kBit/s at a sampling rate of 44.1 kHz and encoded using the AAC codec. All parts of the study were carried out according to the "Wizard Of Oz" method. \\
Twelve participants were recruited from local academic institutions (5 women, 7 men) between the ages of 26 and 47. All participants were students or academic employees in the fields of computer science, architecture, and physics. The study lasted on average 15 minutes and was conducted entirely in English. Five of the twelve participants had prior experience interacting with other robots. Our robot interaction study consisted of two parts that were done entirely during one session. A sketch of the study setup can be seen in Figure \ref{fig:experiment}. In the first part the participants were shown eight different optical illusions, in which two equally valid answers were possible. The robot explained the optical illusion to the participants and determined by means of specific questions which of the possible answers were perceived. For four randomly selected illusions, the robot gave an affirmative response, and subsequently asked the participant whether the alternative illusion was also observed. For the remaining four illusions, the participant was challenged on their answer and the robot proclaimed that the alternative answer was the objectively correct choice, trying to elicit a disagreement response. In the second part, a game called "who am I?" was played with the participant. For this, the participant and the robot took turns choosing one of 16 different objects which were depicted on a sheet of paper and asked each other questions which could be answered with yes or no, to determine the selected object. During the second part, errors in the interaction were simulated at specific points to elicit confusion in the participants. For this purpose, six different interaction errors were designed, from which the experimenter selected one at a time. The interaction errors were: "misunderstanding the participant", "contradicting their own logical reasoning", "choosing an object that didn't exist", "solving the game with the wrong object", "asking the participant questions, instead of answering." and "stopping mid-sentence and repeating question". All interaction errors were at least 30 seconds apart and substantially distinct as to not evoke any anticipation in the participants. Whenever one of the desired user states was visible, the video file was cut into an individual segments with a minimum length of one second and saved with a description file. This included segments, where only the face or the voice displayed one of the user states, which were save separately. All instances of the robot voice were cut from the data, as well as all utterances which primarily amounted to "yes" and "no". In addition, 2.5 minutes of video footage was saved from each participant in which none of the desired user states were visible, as the neutral labeled data. After collecting all the data, the segments were manually annotated by four different annotators. The average inter-annotator agreement (given by Co-hen’s Kappa) was 0.89. Each segment that didn't receive a majority vote for one of the four possible labels was removed. resulting in a total of 415 segments. The total seconds of recorded data used for training can be found in Table \ref{tab:table_groundtruth}.

\begin{figure*}[h]
  \centering
  \includegraphics[width=11cm]{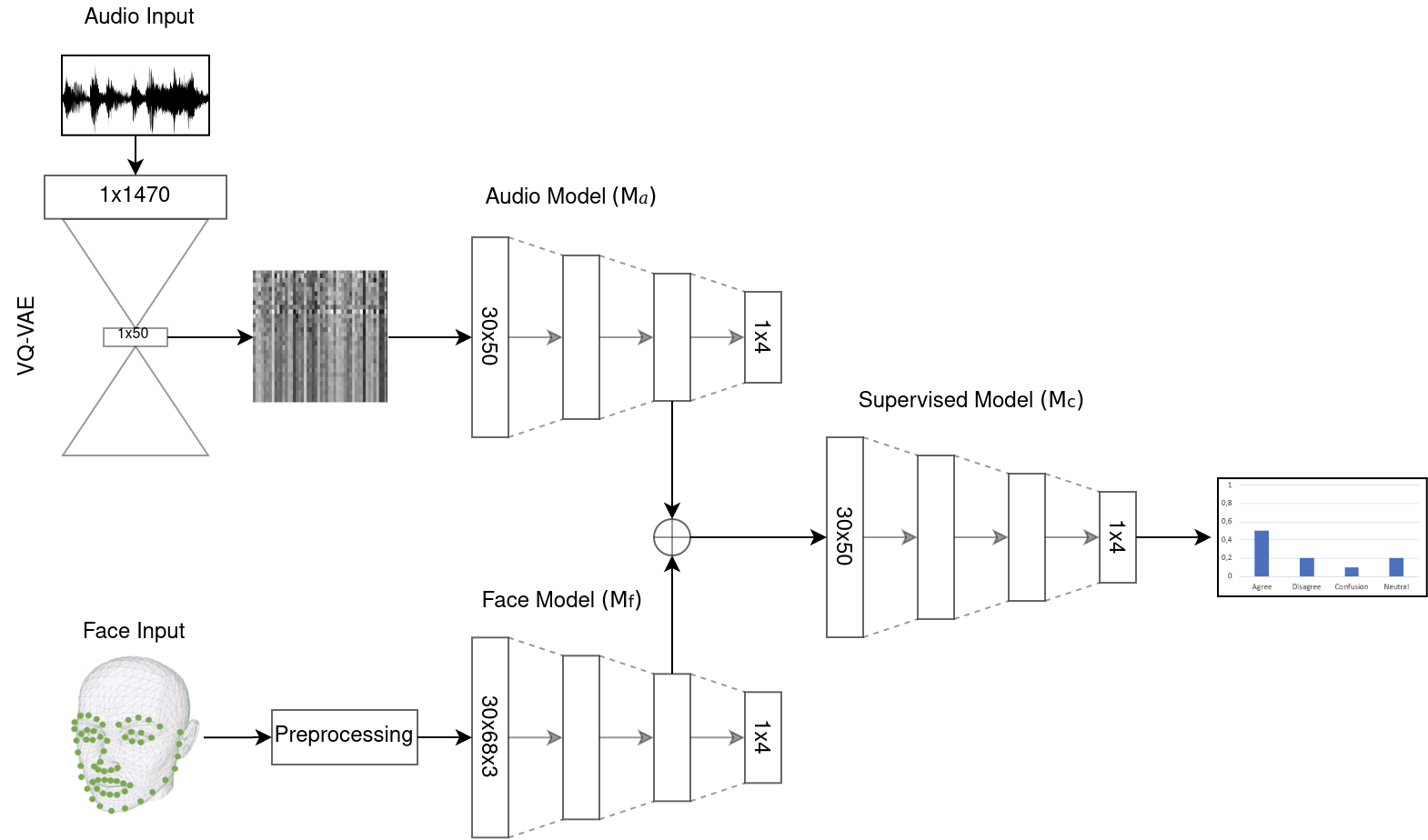}
  \caption{Overview of the model architecture consisting of three different networks, with their individual preprocessing pipelines. The audio model receives the stacked output of the VQ-VAE model, while the Face model receives the preprocessed landmarks. Both the audio model, as well as the face model is trained semi-supervised. The output of all three models is a probability distribution of the four desired classes.}
  \Description[Model Diagram]{An overview of the model architecture. The audio input is fed into the vq-vae model, concatenated and fed into the Audio Model. The face landmarks are concatenated and fed into the face model. The second to last layer of both models are concatenated and become the output of the combined model.}
  \label{fig:diagram}
\end{figure*}

\subsection{Youtube Debate Data Set}
In addition to the labeled data, a large amount of unlabeled data had to be gathered to train the semi-supervised architecture. For this purpose 560 videos of multiple Youtube channels were downloaded in their highest available resolution, at 30fps, as mp4 files \cite{presdebate,JamesCorden,StephenColbert,GrahamNorton,ModernDa92:online}. It was ensured that all videos had at least two different speakers, of whom the head and the upper body were visible. In all videos, the intros were removed, as well as all parts, where a narrator was speaking. Any of the videos that did not have two speakers, where the head was not clearly visible or the head position of one of the speakers was strongly slanted, were removed. The resulting data set contained 210 videos, totaling 230 hours of video footage.

\begin{figure}
  \centering
  \includegraphics[width=6cm]{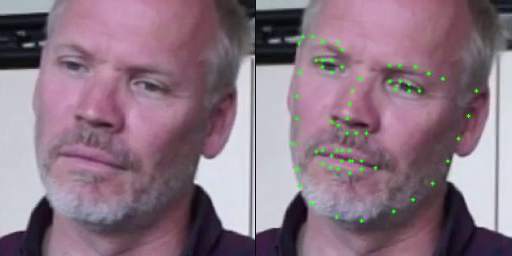}
  \caption{Visualisation of the detected landmarks, from one of the participants of the conducted study, during a confusion event. Left: The clean face without the detected landmarks. Right: the 3D landmarks tracked onto the face.}
  \Description[Visualized Landmark]{A head of a person during a confusion event. The left side shows the person without landmarks in his face and the right side shows the added landmarks}
  \label{fig:landmark_vis}
\end{figure}

\section{Approach}
\label{sec:approach}
The labeled data and the unlabeled data were processed identically. In all videos the faces of the speakers were recognized and 68 3D landmarks, using the model of Bulat \textit{et al.} \cite{bulat2017far}, were added. An example of the landmarks can be found in figure \ref{fig:landmark_vis}. Each face was given a unique ID and tracked for the duration of the video. All videos, in which the audio sampling rate differed were converted to a sampling rate of 44.1 khz using ffmpeg \cite{tomar2006converting}.

\subsection{Audio Pipeline}
Based on the findings of Chorowski \textit{et al.} \cite{Chorowski2019}, we decided against solely encoding the audio data as mel-frequency cepstral coefficients (MFCC) for the training input and instead used a Vector Quantized Variational Autoencoder (VQ-VAE) \cite{Oord2017} with MFCC input and raw data output. We divided the audio data into 1/30 of a second and extracted 18 MFCC features at an input frequency of 44.1 khz, augmented with their temporal first and second derivatives. The encoder consisted of six convolutional layers, with a 2x2 pooling layer after the first, third and fifth layer. The embedding space was defined as a 5x10x1 vector, which was flattened to give a latent variable of 50 values. The decoder consisted of six convolutional layer, with a bilinear upsampling layer after the first, third and fifth layer. After the last convolutional layer, a dense layer was added as the output layer. For all convolutional layer, the Leaky ReLU activation function with an alpha value of 0.3 was used \cite{Maas2013}. We trained the VQ-VAE model on 2000 hours of audio data from randomly selected videos downloaded from the video platform Youtube. Based on the paper by Amari \textit{et al.} \cite{Amari1997AsymptoticST}, we split the data into 1980 hours for the training set and 10 hours each, for the validation and testing set. It was ensured that none of the videos of the Youtube debate data set were part of the data set for the audio pipeline. The training was performed on two NVIDIA TESLA V100 gpu for 72 hours, with each epoch being saved individually. After the training, the epoch with lowest validation loss was restored and used for the preprocessing pipeline.

\subsection{Face Pipeline}
The face pipeline used 30 consecutive frames for each labeled face as input. To minimize the large variance in head movements, we normalized the rotation of the 3D landmarks. To do this, we aligned the average position of all landmarks for each eye along the x-axis, and the average of the eye landmarks with the average of all mouth landmarks, along the y-axis. For the z-axis, we aligned the three foremost landmarks of the nose with the average of all other landmarks. We stored the rotation angles for the first frame in a rotation vector. Starting with the second frame, the rotation vector was subtracted from the current frame and the delta between the current and the last frame was added to the rotation vector. The result were 3D landmarks that were always aligned for the first frame and only changed depending on the movement of the head and face in subsequent frames. For the final preprocessing step, we scaled the 3D landmarks of each frame between zero and one to obtain a uniform input.

\subsection{Model Description}
As shown in Figure \ref{fig:diagram}, we define three different deep learning models, the audio model \(\mathcal{M}_a\), the face model \(\mathcal{M}_f\) and the supervised model \(\mathcal{M}_c\). For \(\mathcal{M}_a\) and \(\mathcal{M}_f\) a wide residual network \cite{Zagoruyko2016} with a depth of 22 and a width of 8, while for \(\mathcal{M}_c\) a depth of 16 and a width of 8 is used. In addition, all residual blocks in the three models are replaced with res2net blocks \cite{gao2019res2net}. For the input of \(\mathcal{M}_a\), 30 consecutive outputs of the VQ-VAE were added together, to form an input vector of 30x50x1, while for \(\mathcal{M}_f\) we use the preprocessed 3d landmarks from the face pipeline, resulting in an input vector of 30x68x3. The model \(\mathcal{M}_c\) receives the concatenated output of the second to last layer from \(\mathcal{M}_a\) and \(\mathcal{M}_f\) as input. All the three models have a fully connected layer with a vector size of 4, using a softmax activation function, as their output layer. An overview of the model architecture can be found in Figure \ref{fig:diagram}.  
For the input, we define three different data sets: audio input only \(x_a\), landmark input only \(x_f\) and combined input \(x_c\), with their respective one-hot label \(y_a\), \(y_f\) and \(y_c\). Additionally, we define the unlabeled audio and landmark data \(u_a\) and \(u_f\). We let \(p_\mathcal{M}(y|x)\) be the predicted class distribution produced by the respective model \(\mathcal{M}\) for the input \(x\) and \(H(p,q)\) be the cross entropy loss, given two probability distributions p and q. We also let \(q_b(\mathcal{M},u)=p_\mathcal{M}(y|u)\) and \(\hat{q}_b(\mathcal{M},u)= arg\: max(q_b(\mathcal{M},u))\).
\\For training \(\mathcal{M}_a\) and \(\mathcal{M}_f\), we use the FixMatch algorithm of Sohn \textit{et al.} with the RandAugment algorithm of Cubuk \textit{et al.} \cite{Sohn2020,Cubuk2020}. We denote the weak and strong augmentations used for the FixMatch algorithm as \(\alpha(.)\) and \(\mathcal{A}(.)\), respectively.  As the input of \(\mathcal{M}_a\) and \(\mathcal{M}_f\) are not natural images we define altered variations of the augmentations used during the semi-supervised training. For \(\alpha(.)\), we only translate the input randomly, by up to 15\%, omitting the random horizontal flip, while for \(\mathcal{A}(.)\) we reduce the set of possible transformations, by removing the shear and rotation transformations. Due to the scarcity of the desired user states inside the unlabeled data sets, we also define a threshold \(\tau_2=\tau^3\), given by the threshold \(\tau\) of the FixMatch algorithm, and a starting epoch \(k\). Starting at epoch \(k\), we use a distilled unlabeled data set given by the output of the respective semi-supervised model, in which any of the first three softmax outputs (agreement/disagreement/confusion) exceeds the threshold \(\tau_2\). Formally, we define \( u_{ad} = \{u \in u_a | max(p_{\mathcal{M}_a}(y|u)) > \tau_2) \wedge arg\, max(p_{\mathcal{M}_a}(y|u))<3\}\) and \( u_{fd} = \{u \in u_f | max(p_{\mathcal{M}_f}(y|u)) > \tau_2) \wedge arg\, max(p_{\mathcal{M}_f}(y|u))<3\}\). Given the epoch \(e\), the batch size \(B\) and the unlabeled batch size factor \(\mu\), the two loss function of the semi-supervised models are given as:

\begin{equation}
\mathcal{L}_{s}(\mathcal{M},y_m,x_m) = \frac{1}{B} \sum\limits_{b=1}^B H(y_{mb},p_{\mathcal{M}}(y|\alpha(x_{mb}))
\end{equation}

\begin{equation}
H_p(\mathcal{M},u_m) = H(\hat{q}_b(\mathcal{M},u_m),p_\mathcal{M}(y|A(u_m)))
\end{equation}
\begin{equation}
\mathcal{L}_{u}(\mathcal{M},u_m) = \frac{1}{\mu B} \sum\limits_{b=1}^{\mu B} \mathbb{1}(max(q_b(\mathcal{M},u_m))H_p(\mathcal{M},u_m)
\end{equation}
During training the two loss functions are combined with the constant hyperparameter \(\beta_1\), which results in the loss functions \(\mathcal{L}_a\) and \(\mathcal{L}_f\), for the semi-supervised models \(\mathcal{M}_a\) and \(\mathcal{M}_f\), respectively.
\begin{equation}
  \mathcal{L}_a =
    \begin{cases}
         \mathcal{L}_{s}(\mathcal{M}_a,y_a,x_a) + \beta_1\mathcal{L}_{u}(\mathcal{M}_a,u_a) & e < k\\
         \mathcal{L}_{s}(\mathcal{M}_a,y_a,x_a) + \beta_1\mathcal{L}_{u}(\mathcal{M}_a,u_{ad}) & e >= k \\
    \end{cases}          
\end{equation}

\begin{equation}
  \mathcal{L}_f =
    \begin{cases}
         \mathcal{L}_{s}(\mathcal{M}_f,y_f,x_f) + \beta_1\mathcal{L}_{u}(\mathcal{M}_f,u_f) & e < k\\
         \mathcal{L}_{s}(\mathcal{M}_f,y_f,x_f) + \beta_1\mathcal{L}_{u}(\mathcal{M}_f,u_{fd}) & e >= k \\
    \end{cases}          
\end{equation}
The loss function of the model \(\mathcal{M}_c\) is the cross-entropy loss, given by \(\mathcal{L}_{s}\):

\begin{equation}
\mathcal{L}_{c} = \mathcal{L}_{s}(\mathcal{M}_c,y_c,x_c)
\end{equation}
We combine all three loss functions, with two constant hyperparameters \(\beta_2\) and \(\beta_3\), to receive the final loss function
\begin{equation}
\mathcal{L} = \beta_2(\mathcal{L}_a + \mathcal{L}_f) + \beta_3\mathcal{L}_c
\end{equation}
During the training we minimize the loss function \(\mathcal{L}\) by performing three consecutive forwards steps through the model, followed by the weighted backpropagation step. For the \(\mathcal{M}_a\) and \(\mathcal{M}_f\) networks, we take a batch of size \(B*\mu\) from the distilled unlabeled Youtube data set and perform the pseudo labeling and consistency regularization step on them. Together with a labelled batch of size \(B\) from \(x_a\) and \(x_f\), we calculate \(\mathcal{L}_f\) and \(\mathcal{L}_a\) on their respective 1x4 dense layers. For the network \(\mathcal{M}_c\) we take a batch of size \(B\) from the data set \(x_c\) and calculate the cross-entropy loss \(\mathcal{L}_{c}\). With the loss from \(\mathcal{M}_c\), \(\mathcal{M}_a\) and \(\mathcal{M}_f\) we calculate the final loss \(\mathcal{L}\) and perform the backpropagation step over the whole network. We use the hyperparameters \(\beta_2\) and \(\beta_3\) to balance the influence of the supervised and semi-supervised training.   


\begin{table}
 \caption{We compare the F1-score of both semi-supervised models, trained in isolation, against the F1-score of the model, that combines both modalities. All F1-scores are depicted as mean score and standard deviation, with the highest F1-score in bold\\}
  \centering
  \resizebox{\columnwidth}{!}{
  \begin{tabular}{lllll}
    \toprule
    Class  & Audio only & Face only & Combined      \\
    \midrule
    Agreement & 62.6\(\pm2.94\)\% & 76.2\(\pm0.72\)\% & \textbf{86.7\(\pm1.13\)\%}    \\
    Disagreement     & 46.3\(\pm3.99\)\% & 75.1\(\pm1.21\)\% & \textbf{75.5\(\pm1.45\)\%}   \\
    Confusion     & 32.3\(\pm3.14\)\% & 47.6\(\pm1.13\)\% & \textbf{57.1\(\pm1.62\)\%}         \\
    Neutral     & 29.1\(\pm2.53\)\% & 44.9\(\pm0.87\)\% & \textbf{46.3\(\pm1.51\)\%}         \\
    \bottomrule
  \end{tabular}}
  \label{tab:table_result}
\end{table}
\begin{table}
 \caption{The F1-score of all three models trained completely supervised, against the F1-score of the semi-supervised approach. All F1-scores are depicted as mean score and standard deviation, with the highest F1-score in bold.}
  \centering
  \resizebox{\columnwidth}{!}{
  \begin{tabular}{lll}
  
    \toprule
    Class  & Supervised & Semi-Supervised      \\
    \midrule
    Agreement    & 79.5\(\pm3.63\)\% & \textbf{86.7\(\pm1.13\)\%}    \\
    Disagreement &74.2\(\pm3.21\)\% &  \textbf{75.5\(\pm1.45\)\%}  \\
    Confusion    & 48.9\(\pm2.49\)\% & \textbf{57.1\(\pm1.62\)\%}          \\
    Neutral      & 35.5\(\pm2.29\)\% & \textbf{46.3\(\pm1.51\)\%}          \\
    \bottomrule
    
  \end{tabular}}
  \label{tab:table_semisuper}
\end{table}

\begin{figure}[h]
  \centering
  \includegraphics[width=8cm]{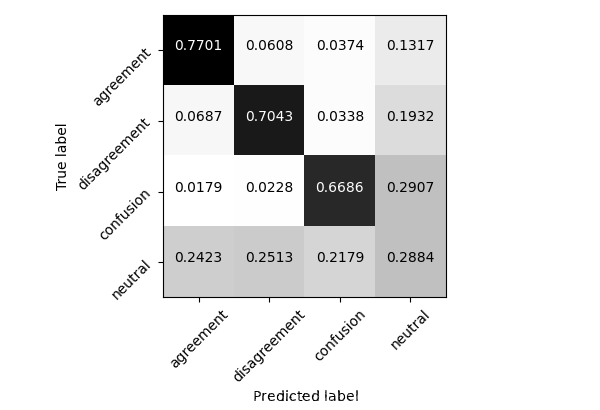}
  \caption{The confusion matrix visualising the classification of the test set for the supervised model. Each true label is normalized over their respective row.}
  \Description[Supervised Confusion matrix]{The confusion matrix of the supervised model, which shows an accuracy of 77\% for agreement, 70\% for disagreement, 67\% for confusion and 29\% for neutral.}
  \label{fig:confusion_super}
\end{figure}

\begin{figure}[h]
  \centering
  \includegraphics[width=8cm]{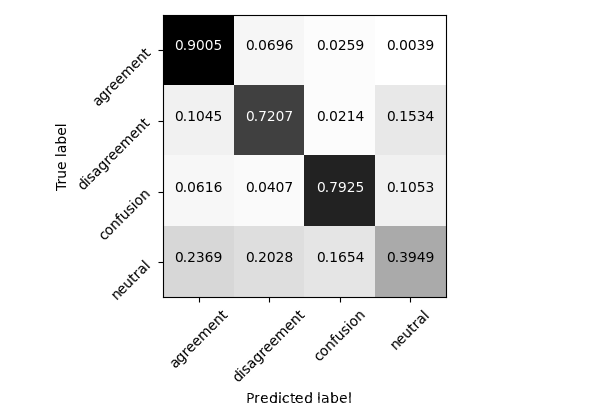}
  \caption{The confusion matrix visualising the classification of the test set for the semi-supervised model. Each true label is normalized over their respective row.}
  \Description[Semi-Supervised Confusion matrix]{The confusion matrix of the semi-supervised model, which shows an accuracy of 90\% for agreement, 72\% for disagreement, 79\% for confusion and 39\% for neutral.}
  \label{fig:confusion_semisuper}
\end{figure}

\begin{table*}[tb]
            \centering
            \resizebox{\textwidth}{!}{
            \begin{tabular}{c|c|c|c|c|c|c}
                  &   &   &   & \multicolumn{3}{|c}{F1 score}\\
                Method & sata set & Features & Data & Agreement & Disagreement & Confusion\\
                \hline
                 Hahn et al. (2006) \cite{Hahn2006} & ICSI data set & Audio Features & 1.800 segments & \multicolumn{2}{|c|}{75.0\%} & - \\
                 Germesin and Wilson (2009) \cite{Germesin2009} & AMI Meeting Corpus & Audio Features & 19.600 segments & 45.2\% & - & - \\
                 Wang et al. (2011) \cite{Wang2011} & DARPA GALE & Audio Features & 2.589 utterances & 57.2\% & 51.2\% & - \\
                 Bousmalis et al. (2011) \cite{Bousmalis2011} & Canal9 data set & Face+Audio+Body Features & 147 episodes & \multicolumn{2}{|c|}{accuracy: 64.2\%} & - \\
                 Rosenthal et al. (2015) \cite{rosenthal-mckeown-2015-couldnt} & ABCD data set & Text Features & 200,000 posts & 58.5\% & 73.0\% & - \\
                 Hiray et al. (2018) \cite{Hiray2018} & ABCD data set & Text Features & 200,000 posts & \multicolumn{2}{|c|}{80.0\%} & - \\
                 Yang et al. (2016) \cite{Yang2016} & own data set & Brain wave + Audio Features & 186 minutes & - & - & \textbf{87.8\%} \\
                 Ni et al (2017) \cite{ni2017confused} & Kaggle data set  & Brain wave data & 100 videos & - & - & 73.3\% \\
                 Borges et al. (2019) \cite{Borges2019} & own data set  & Face Action Units & 8.160 segments & - & - & 80.9\% \\
                 ours (supervised) & own data set  & Face+Audio Features & 415 segments & 79.5\% & 74.2\% & 48.9\% \\
                 ours (semi-supervised) & own data set  & Face+Audio Features & 415 segments & \textbf{86.7\%} & \textbf{75.5\%} & 57.1\% \\

            \end{tabular}
            
     }
    \caption{Summary of existing methods that have attempted dis-/agreement and confusion classification with their respective F1 score, if available. As the table shows multiple different modalities, the comparability between the approaches should be treated with caution. The best score for each category are marked in bold.}
   \label{table_summary}
\end{table*}

\section{Results and Discussion}
\label{sec:results}
All training was done using the hyperparameters: \(B=12\), \(k=10\), \(e=500\), \(\beta_1=1\), \(\mu=10\), \(\beta_2=3\) and \(\beta_3=2\). The hyperparameters \(B, k, \beta_1, \beta_2,\) and \(\beta_3\) were determined using the Hyperopt python package \cite{bergstra2013making} with a maximum evaluation number of 100. We trained the model on two NVIDIA TESLA V100 with a five fold cross-validation, in which one of the five sets was used for validation, while the other four were used for training and report the average F1-score of the trained models in Table \ref{tab:table_result}. Due to the semi-supervised learning our model has three distinct outputs, the audio only output, the face only output and the combined modality output. we report the average F1-score of the cross-validation for both the single modality models and the combined model, to asses the classification quality of the different models. Comparing the single modality models with the full model shows an increase in F1-score for all four classes when the multimodal approach is used, indicating that the combined model can incorporate the input data from both single modality models successfully.
It can be seen that the audio-only model has the lowest F1-score in all classes and exhibits a higher standard deviation compared to the other models which is expected, as audio data contains more voice variance, due to different voice profiles, and general noise than the face data. In addition, the variance of the model can be a byproduct of the high compression performed by the VQ-VAE. Although the VQ-VAE can reconstruct clearly understandable voice samples, some features could have been lost as the latent output of the model is comparatively small, with only two percent of the input data.
In terms of individual classes, agreement shows the largest increase over the average score for each individual modality at 17.3\%, while disagreement and confusion exhibit marginally smaller increases at 14.80\% and 17.15\%, respectively. 
It is noteworthy that the F1-score of the disagree and neutral classes for the multimodal approach show only a small increase in their respective F1-score of 0.4\% and 1.4\% compared to the face-only model, whereas the agree and confusion classes show a considerably higher increase with 10.5\% and 9.5\%, respectively. While it was expected that agreement would benefit strongly from the multi model approach, due to the high correlation of features \cite{Bousmalis2011}, the comparatively low increase of the F1-score for disagreement was unexpected.\\
For the purpose of investigating the influence of the semi-supervised trained models on the performance of the combined model, we trained all models in a fully supervised manner. A cross-entropy loss was used for all models with the same five fold cross-validation. As can be seen in Table \ref{tab:table_semisuper}, all four classes have an increase in their F1-score, with neutral exhibiting the strongest increase with 10.8\%, which shows that the semi-supervised training does have an effect on the performance of the trained models. The high performance gain for the confusion class also shows that complex emotions, which are normally difficult to detect, can benefit greatly from training with unlabeled training data. 
To compare the classification errors for each class, we created a confusion matrix for the full semi-supervised model, as well as the supervised model. As can be seen in Figure \ref{fig:confusion_super}, the supervised model has a very high false positive, as well as a high false negative rate, while the confusion matrix for the semi-supervised model in figure \ref{fig:confusion_semisuper} shows a far higher false negative rate than false positive rate. This would suggest, that the semi-supervised model is able to more accurately detect events that it has already seen before than the supervised model, while failing to detect events that differ strongly from the events in the training data. This was expected, as the FixMatch algorithm uses strongly distorted input to enrich the given data, which can increase the robustness of the model. \\
As can be seen in table \ref{table_summary} our multimodal semi-supervised approach outperforms other models in regards to dis-/agreement classification by up to 6.7\%. As we only used 5\% of the training data compared to Borges \textit{et al.} \cite{Borges2019}, we were unable to capture all distinct characteristics of confusion and therefore didn't reach a comparatively high accuracy for the confusion detection.   
As our networks were mostly trained on footage with a still camera, we found that the networks performed similarly well when there was only slight camera movement, but degraded when there was a cut between the cameras or large movements in the camera positions. As currently the preprocessing of the face movement doesn't take the camera movement into account, this could be alleviated by subtracting the camera movement from the face movement.

\section{Conclusion}
In this work, we developed and evaluated a method for detecting mental states of users during human-robot interactions, using only a small set of labeled data. We found that semi-supervised networks can extract meaningful user state representations using a small set of labeled data for one modality, which in turn can help to increase the overall performance of a multimodal supervised network when faced with a data scarcity problem. We also showed that modern semi-supervised approaches not only work for image recognition tasks, but also for tasks involving social signals, especially complex social signals such as confusion, increasing the accuracy and robustness of the model. Further research is needed to determine the performance of our approach when different data set are utilized or different unlabeled data is used during training. In the future, we plan and extend our work to a larger number of different modalities and further research the achieved robustness of the model.
\label{sec:conclusion}

\bibliographystyle{ACM-Reference-Format}
\bibliography{icmi}

\appendix

\end{document}